%% file: ms.tex
\documentclass[twocolumn,10pt]{asme2ej}

\usepackage[captionskip=2pt]{floatrow}
\newlength\Myfigwd

\usepackage{hyperref}

\newcommand{\realfield}[1]{\hbox{I \kern -.25em R}^{#1}}   
\newcommand {\mb}[1]{\mathbf{#1}}
\newcommand {\bs}[1]{\boldsymbol{#1}}
\newcommand{\uvec}[1]{\hat{\mathbf{#1}}}

\newcommand{\T}{^{\mathrm{T}}}  
\newcommand{\rmd}{\textrm{d}}  

\usepackage[%
font=small,
labelfont=bf,
singlelinecheck=off]{caption}
\usepackage{array}
\newcolumntype{P}[1]{>{\centering\arraybackslash}p{#1}}
\newcolumntype{M}[1]{>{\centering\arraybackslash}m{#1}}

\makeatletter
\newcommand{\thickhline}{%
    \noalign {\ifnum 0=`}\fi \hrule height 1pt
    \futurelet \reserved@a \@xhline
}
\newcolumntype{"}{@{\hskip\tabcolsep\vrule width 1pt\hskip\tabcolsep}}
\makeatother

\usepackage{lipsum}
\usepackage{graphicx}
\usepackage{comment}
\usepackage{gensymb}
\usepackage{amsmath}
\usepackage{amssymb}
\usepackage{mathtools}
\usepackage{cancel}
\usepackage{commath}
\usepackage{etoolbox}
\usepackage{caption}
\usepackage{subcaption}
\usepackage{siunitx}
\usepackage[compress]{cite}
\usepackage{subcaption}
\renewcommand{\citedash}{--}
\usepackage{float}
\floatstyle{plaintop}
\restylefloat{table}
\usepackage{enumitem}
\usepackage{arydshln}
\usepackage{soul}
\usepackage{longtable}
\usepackage{colortbl}
\usepackage[table,xcdraw]{xcolor}
\usepackage{tabu}
\usepackage{caption}

\graphicspath{
	{figures/png/}
	{figures/pdf_fig/}}

\begin{document}
\input{content/title-authors}
\maketitle    

\input{content/abstract}
\input{content/asme_nomenclature}

\input{content/intro}
\input{content/related_work}
\input{content/mechanism}
\input{content/kinematics}

\input{content/imu_force_estimation}
\input{content/experiments}
\input{content/conclusion}

\bibliographystyle{asmems4}
\bibliography{%
	bib/continuum_robot_general,%
	bib/aerial_manipulation,%
	bib/aerial_manipulation_compliant,%
	bib/justin_paper,%
	bib/long_added_references,%
	bib/kinematics}

\end{document}

%% file: content/title-authors.tex
\author{Guoqing Zhang\thanks{Equal contribution}\\[2pt]
	{\tensfb Qianwen Zhao\footnotemark[1]}\\[2pt]
	{\tensfb Long Wang}\thanks{Address all correspondence for other issues to this author.}\\[2pt]
	\affiliation{Department of Mechanical Engineering\\
		Schaefer School of Engineering \& Science\\
		Stevens Institute of Technology \\
		Hoboken, NJ, 07030 USA\\
		gzhang21@stevens.edu, qzhao10@stevens.edu, lwang4@stevens.edu
	}
}

\title{A Lightweight Modular Continuum Manipulator with IMU-based Force Estimation}

%% file: content/abstract.tex
\begin{abstract}
{\it Most aerial manipulators use serial rigid-link designs, which results in large forces when initiating contacts during manipulation and could cause flight stability difficulty. This limitation could potentially be improved by the compliance of continuum manipulators.\par
To achieve this goal, we present the novel design of a compact, lightweight, and modular cable-driven continuum manipulator for aerial drones. We then derive a complete modeling framework for its kinematics, statics, and stiffness (compliance). The modeling framework can guide the control and design problems to integrate the manipulator to aerial drones. 
In addition, thanks to the derived stiffness (compliance) matrix, and using a low-cost IMU sensor to capture deformation angles, we present a simple method to estimate manipulation force at the tip of the manipulator.
We report preliminary experimental validations of the hardware prototype, providing insights on its manipulation feasibility. We also report preliminary results of the IMU-based force estimation method.
}
\end{abstract}

%% file: content/asme_nomenclature.tex
\begin{nomenclature}
	\small
	\entry{Frame \{B\}}{The base frame with $\mb{b}$ located at the center of the base, $\uvec{x}_b$ passing through the first wire and $\uvec{z}_b$ perpendicular to the base.}
	\entry{Frame \{1\}}{Characterizes the plane in which the continuum segment bends and it is obtained by a rotation of $\delta$ about $\uvec{z}_b$. Unit vector $\uvec{x}_1$ is along the projection of the central axis on the plane of the base and $\uvec{z}_1=\uvec{z}_b$.}
	\entry{Frame \{E\}}{It is defined with $\uvec{z}_e$ as the normal to the end surface and $\uvec{x}_e$ is the intersection of the bending plane and the end disk top surface.}
	\entry{Frame \{G\}}{Frame \{G\} defined with $\uvec{z}_g=\uvec{z}_e$ and its $\uvec{x}_g$ passing through the $1^\text{st}$ wire. It can be obtained by a rotation angle $\delta_e$ about $\uvec{z}_e$ which is the unit vector normal to the end disk. This angle is given by $\delta_e = -\delta$.}
	\entry{$i$}{Index of the tendons, $i$=1,2,3,4.}
	\entry{$r$}{Radius of the pitch circle where the tendons are distributed.}
	\entry{$\delta$}{Bending plane angle.}
	\entry{$\theta(s)$}{Bending angle; the angle of the $\uvec{z}_1$ or $\uvec{z}_b$ to the tangent at the s arc length position in the bending plane. $\theta(L)=\theta$.}
	\entry{$L,L_{i}$}{Length of central backbone, Length of $i^{th}$ tendon measured from the base disk to the end disk.}
	\entry{$q_{i}$}{The displacement of $i^{th}$ tendon.}
	\entry{$\beta$}{Division angle of the tendons along the circumference of the pitch circle, $\beta$=$\pi$/2.}
	\entry{${ }^{a} \mathbf{T}_{b}$}{Homogeneous transformation matrix describing orientation and position of frame\{b\} with respect to frame\{a\}.}
	\entry{${ }^{a} \mathbf{p}_{b},^{a} \mathbf{R}_{b}$}{Position vector and rotation matrix of frame\{b\} with respect to frame\{a\}, respectively.}
	\entry{$\psi,\dot{\psi}$}{Configuration space vector and corresponding time derivative,$\psi = [\theta, \delta]^{\mathrm{T}}$.}
	\entry{$\mb{q},\dot{\mb{q}}$}{Tendon displacement vector and corresponding time derivative,$\mathbf{q} = [ q_1, q_2, q_3, q_4]^{\mathrm{T}}$.}
	\entry{$\dot{\mb{x}}$}{ vector in task space.}
	\entry{$\mb{J}_{\mathbf{q}\psi}$}{A Jacobian matrix mapping the configuration space velocities to joint velocities.}
	\entry{$\mb{J}_{\mathbf{\mb{x}}\psi}$}{A Jacobian matrix mapping the configuration space velocities to task space velocities.}
	\entry{$\mb{J}_{\mathbf{v}\psi}$}{A Jacobian matrix mapping the configuration space velocities to linear velocities.}
	\entry{$\mb{J}_{\mathbf{\omega}\psi}$}{A Jacobian matrix mapping the configuration space velocities to angular velocities.}
	\entry{$\mb{E},\nabla \mb{E}$}{Potential elastic energy of the continuum arm and corresponding gradient with respect to the configuration space}
	\entry{$\bs{\tau}$}{Tendon actuation force, $\bs{\tau}=\left[\tau_{1}, \ldots, \tau_{4}\right]^{\mathrm{T}}$.}
	\entry{$E_{p}, E_{T}$}{Young's modulus of the central NiTi backbone and Young's modulus of the tendon, respectively.}
	\entry{$I_{p}$}{Cross-sectional second moment of inertia of the NiTi backbone.}
	\entry{$A$}{Cross-sectional area of the tendon.}
	\entry{$\mb{w}_{\text{ext}},\mb{F}_{\text{ext}}$}{External wrench and force applied to the end disk tip of the continuum arm, respectively.}
	\entry{$\mb{K}_X,\mb{K}_{\psi}$}{Task space stiffness and configuration space stiffness, respectively.}
	\entry{$\mb{K}_q$}{Stiffness matrix of tendons}
	\entry{$\mb{F}^{*}$}{Generalized force applied to continuum arm in configuration space}
\end{nomenclature}

%% file: content/intro.tex
\section{Introduction}
\label{section:intro}

According to the Federal Aviation Administration (FAA) aerospace forecast release \cite{FAA_Forecast}, there were approximately 1.1 million unmanned aerial vehicles (UAVs), or so-called drones, in the United States as of December 31, 2017. Today, multirotor UAVs are household products and are widely used for photography, mapping and inspection, or simply for recreation. Despite their success in perceptual tasks, the mainstream uses of UAVs do not involve physical interactions with objects or the environment. 
The ability to physically manipulate objects and environments unlocks new tasks that aerial drones could be capable of. To this end, many researchers have spent great efforts to enhance aerial drones with manipulation capabilities \cite{Pounds2011,Mellinger2011,Korpela2013,Scholten2013,Korpela2014,Garimella2015,Anzai2018,Nguyen2018,Ruggiero2018,Baraban2020,Kim2020}, and the investigations span a variety of topics such as aerial grasping, aerial interaction force, flight and payload dynamics, cooperative control, model-predictive control, and haptics-based manipulation. \par
Compliance has been introduced to manipulation tasks as an advantage because it can potentially provide robustness to errors. In the case of aerial manipulation, compliance can also help increase the safety margin by softening the collision impact when initiating contacts during manipulation. Recent works have explored the use of compliance in aerial manipulation using soft manipulators, compliant manipulators, and underactuated hands \cite{Pounds2011,Yuksel2015,Suarez2015,Suarez2018,Samadikhoshkho2020,Fishman2021,Fishman2021a}. \par
\begin{figure}[!t]
	\centering
	\includegraphics[width=0.8\columnwidth]{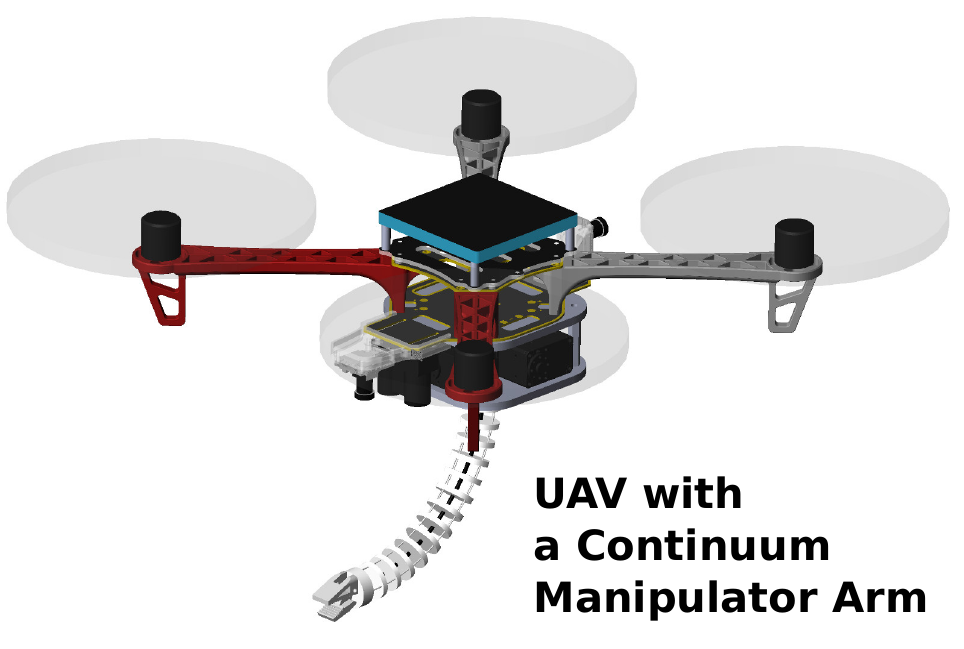}
	\caption{Schematic of an unmanned aerial vehicle(UAV) integrated with a continuum manipulator arm}
	\label{fig:schematic of overall system}
\end{figure}

Continuum manipulators are an unconventional class of robots having inherited compliance in their bio-inspired structures such as an elephant's trunk or an octopus's tentacles \cite{Hannan2003,Liu2019,Yeshmukhametov2019,Zheng2012,Zheng2013,Cianchetti2012,Cianchetti2015,Gravagne2002,Gravagne2002a,Simaan2004,Conrad2015,Conrad2017,simaan2018medical,McMahan2006,Rone2018,Rone2019,sitler2022modular}. In this work, we introduce continuum manipulators to aerial manipulation, taking advantage of the mechanism compliance, and the compact and lightweight design. Figure~\ref{fig:schematic of overall system} illustrates our vision for using modular lightweight continuum manipulators in aerial manipulation.

This work was initially presented in \cite{zhao2022modular}. Compared to our prior work in \cite{zhao2022modular}, this work presents the same three key contributions while extending the work to have two additional key contributions. The three contributions presented in \cite{zhao2022modular} include the following:
\begin{enumerate}
	\item We present the hardware design of a compact, lightweight, and modular cable-driven continuum manipulator for aerial drones.
	\item We derive a complete modeling framework for kinematics, statics, and stiffness (compliance) matrix.
	\item We report preliminary experimental validations of the hardware prototype, providing insights on its integration feasibility.
\end{enumerate}\par
The two additional contributions (since \cite{zhao2022modular} being published) include: 
\begin{enumerate}
    \setcounter{enumi}{3}
    \item We present a low-cost sensor integration using Inertia Measurement Units (IMUs).
    \item We develop and validate an IMU-based force estimation framework that utilizes the IMU feedback and the derived stiffness (compliance) matrix.
\end{enumerate}

%% file: content/related_work.tex
\section{Related Work}
\subsection{Aerial Manipulation.}\quad  Pounds \textit{et~al}. \cite{Pounds2011} and Mellinger \textit{et~al}. \cite{Mellinger2011} investigated aerial grasping control methods considering payload dynamics and demonstrated on a helicopter platform and on a quadrotors, respectively.
Scholten \textit{et~al}. presented interaction force control of an aerial drone equipped with a Delta parallel manipulator\cite{Scholten2013}. 
Korpela \textit{et~al}. investigated dynamic stability of a mobile UAV equipped with two 4-DoF manipulators in \cite{Korpela2013} and continued to develop a framework for the valve turning task in \cite{Korpela2014}. 
Nguyen \textit{et~al}. presented a novel method for aerial manipulation where multiple quadrotors are connected via passive spherical joints and generate thrusts for manipulation tasks collectively \cite{Nguyen2018}.
Anzai \textit{et~al}. developed a transformable multirotor robot with closed-loop multilinks structure that can adapt to shapes of large objects to be grasped \cite{Anzai2018}.
Garimella and Kobilarov introduced model-predictive control in aerial pick-and-place tasks \cite{Garimella2015}, and Baraban \textit{et~al}. continued to explore an adaptive parameter estimation method for unknown mass of objects \cite{Baraban2020}.
Kim and Oh developed a haptics-based testing-and-evaluation platform for drone-based aerial manipulation \cite{Kim2020}.\par

\subsection{Aerial Manipulation with Compliance.}\quad A compliant underactuated manipulator was used in Pounds \textit{et~al}.'s work \cite{Pounds2011}. 
Suarez \textit{et~al.} developed a lightweight compliant arm for aerial drones and proposed its usage in payload mass estimation\cite{Suarez2015}. The same authors extended the compliant arm design for dual-arm aerial manipulation \cite{Suarez2018}. 
Y\"uksel \textit{et~al.} presented another lightweight compliant arm design using a cable-driven flexible-joint for aerial manipulation \cite{Yuksel2015}. 
Samadikhoshkho \textit{et~al.} introduced a modeling framework for considering the use of continuum manipulators for aerial manipulation tasks and reported findings in a simulated environment \cite{Samadikhoshkho2020}.
Fishman \textit{et~al.} prototyped a soft gripper for aerial manipulation, developed algorithms for trajectory optimization considering the soft gripper modeling, and demonstrated aerial grasping in both simulations and physical experiments \cite{Fishman2021,Fishman2021a}.
\par

\subsection{Continuum Manipulators.}\quad Continuum manipulators have been widely used in biomedical applications \cite{Simaan2004, Conrad2015, Conrad2017, simaan2018medical} because of their flexibility and compliance, enabling safe interaction with the anatomy. They are also good candidates for field robotics applications.
McMahan \textit{et~al.} demonstrated the field capabilities and manipulation of a pneumatic continuum manipulator attached to a mobile robot platform in \cite{McMahan2006}. 
Rone \textit{et~al.} used a continuum tail to stabilize a walking mobile robot \cite{Rone2018,Rone2019}.
Sitler and Wang recently presented a modular continuum manipulator design for free-floating underwater manipulation \cite{sitler2022modular}.

\subsection{Continuum Robot Force Sensing via Estimating Shape Change}
When exerting an external force, due to the compliance, continuum robots exhibit a deformation in their shape or tip pose. Researchers have taken advantage of this behavior for estimating force using a variaty of sensory modalities that can detect shape deformation. We draw inspirations from the works below to develop an IMU-based force estimation method in this work.\par 
Bajo and Simaan demonstrated contact detection by capturing the instantaneous screw axis via Electromagnetic (EM) sensors \cite{bajo2011kinematics}, followed by Chen \textit{et~al.} adopting a similar approach for soft robots \cite{chen2020modal}. Back \textit{et~al.} proposed a force estimation method for steerable catheters through bi-point tracking \cite{back2018three}. Khoshnam \textit{et~al.} developed a method for estimating tip contact force for steerable catheters using bending shape curvatures \cite{khoshnam2015modeling}. Qiao \textit{et~al.} presented force estimation from shape using fiber Bragg grating (FBG) sensors \cite{qiao2021force}. 

%% file: content/mechanism.tex
\section{Mechanism Description}
The mechanical structure of the proposed design comprises of the continuum arm portion and the driving mechanism portion. In this section, we start with a discussion of three design considerations, and followed by detailed descriptions of the two parts individually. 

\subsection{Design Considerations}

The first design decision was to determine what type of continuum arm mechanism to use. The mechanical structure of a continuum robot can be categorized into a soft structure continuum arm \cite{JONES2004,Fras2018} or a multi-disk continuum arm with single or multi segment(s) \cite{Xu2021,Neumann2016}. Among many different driving mechanisms, the most commonly used ones are motor-actuated tendon driven robots \cite{Xu2021,Renda_2012,Neumann2016} and soft structures pressurized by pneumatic actuators \cite{JONES2004,McMahan_2005}. For our aerial manipulation application, we value the arm driving accuracy and the kinematic computation simplicity the most. Therefore, a tendon-driven, multi-disk continuum robot design was selected. \par

The next important feature is lightweight, due to UAV's payload limitation. Relatively small and compact servo motors were used, and detailed implementation is discussed in Section~\ref{sec:actuation unit}. Most components of the design are 3D printed with Ploylactic Acid (PLA). The lightweight compliant structure is stiffened by using NiTi cables as the continuum backbones. \par

The final design feature is modularization. Specifically, (i) the design should be adaptable to various drone platforms, and (ii) the arm length can be adjusted for different task requirements. In achieving (i), the UAV adapter of the aerial continuum robot needs to be customizable. For (ii), the stackable modular design allows for adding additional arm segments. \par

\begin{figure}[!h]
    \centering
    \includegraphics[width=0.97\columnwidth]{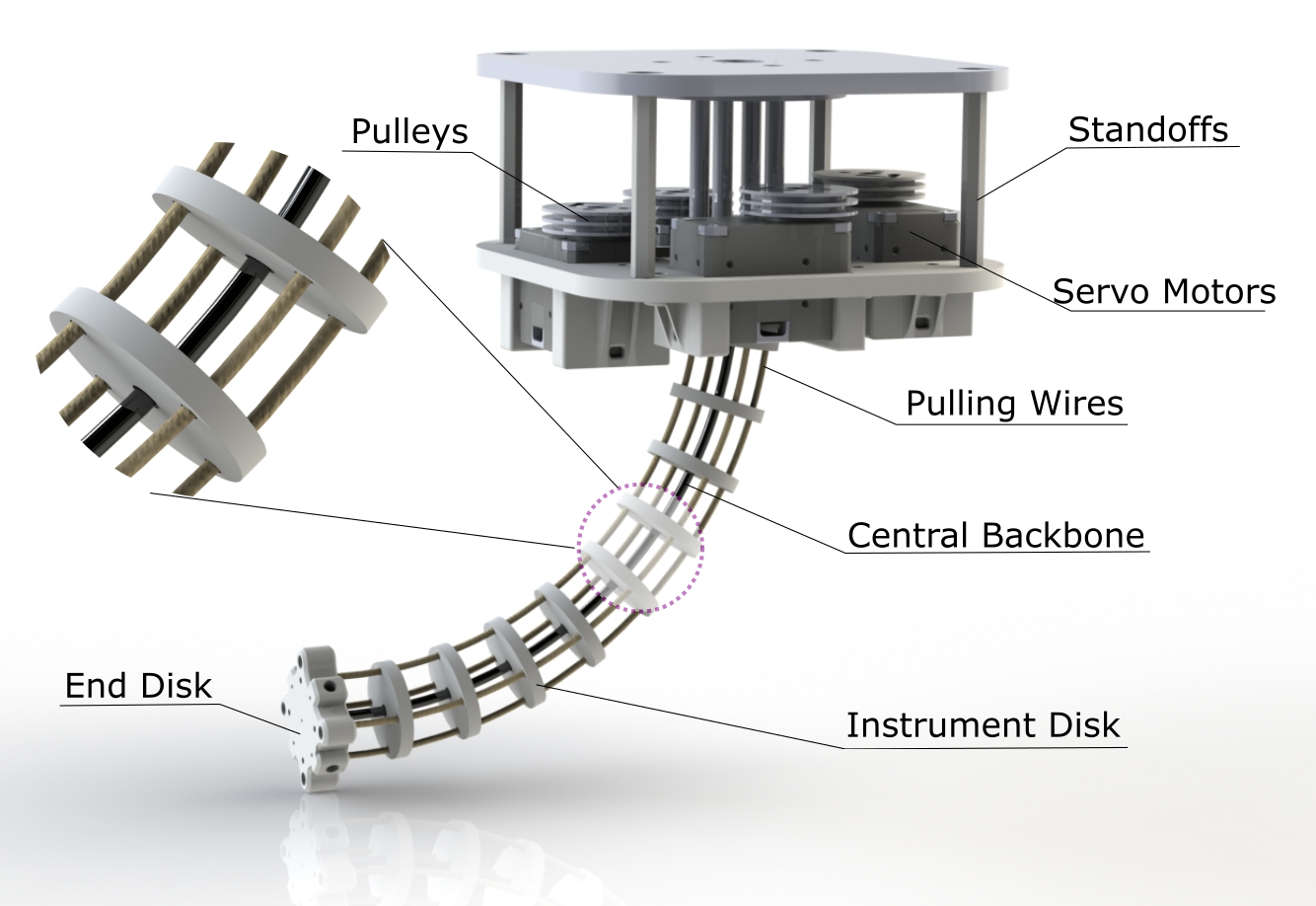}
    \caption{Full Assembly of the proposed modular continuum arm and a close-up view: primary backbone is in black and four tendons are in beige.}
    \label{fig:fullassembly}
\end{figure}

\subsection{Continuum Arm Design}

Figure~\ref{fig:fullassembly} depicts the proposed modular continuum arm for aerial manipulation. The top portion is the driving mechanism (or actuation unit, Section \ref{sec:actuation unit}), and the lower-half illustrates the continuum arm design. The arm specifications can be found in Table~\ref{tab:simulation}.\par

\input{content/tab_parameter}

The arm consists of a central backbone, four tendons, nine circular structural disks, an end disk, and a base disk integrated with a cable routing mechanism. The central backbone is secured at the base and the end disk to keep it at a constant length while bending. Four tendons are fastened at the end disk as well, which allows the arm to articulate. Nine equally distributed circular disks regulate the four pulling tendons aligned with the central backbone while pulling/releasing. A closer inspection of the tendon distribution can be found in the top left corner of Fig.~\ref{fig:fullassembly}. The arm prototype being built is 420 grams in total, including both the continuum arm design and the actuation unit. It is feasible for a relatively lightweight commercial drones to carry. \par

To minimize the variation of mechanical properties between arm prototypes, the group introduced 3D printed spacers to keep the circular disks evenly distributed. Such improved arm prototype was utilized in the external force estimation experiment (Section \ref{sec:force est exp}).

\subsection{Actuation Unit}\label{sec:actuation unit}

The actuation unit consists of 4 low-cost lightweight servo motors (Dynamixel XL430-W250-T, 57.2g each) and a tendon routing mechanism. The servo motors are driven by a custom motor shield and a Teensy 4.0 as the main micro-controller due to its fast processing speed and compact form. Each tendon (fishing wire) is passed from the arm to the actuation unit through a vertically installed bearing and is wrapped on a motor pulley. During arm articulaton, certain motor(s) draw wires while their opposites release wires, and such coordination drives the end-effector to its designated pose. Detailed motor coordination and kinematics modeling can be found in the next section. 

%% file: content/tab_parameter.tex
\begin{table}[!t]
\small
\caption{Parameters of the aerial continuum robot}
\begin{tabular}{m{0.25\columnwidth}m{0.3\columnwidth}m{0.25\columnwidth}}
 \hline \\
$L= 222 \ mm$                   & $r= 12 \ mm$                      & $E_p= 82\ GPa $                                                 \\ \hline\\
$E_T= 2.34\ GPa$                     & $I_p= 0.2485\ mm^4 $                    & $A = 0.2642\ mm^2$                                                     \\ \hline
\end{tabular}
\label{tab:simulation}
\end{table}
%

%% file: content/kinematics.tex
\section{Kinematics, Statics and Stiffness Modeling}
The nomenclature described earlier is referenced to facilitate the modeling of kinematics, mechanics and stiffness of the continuum structure.  \par
\subsection{Kinematics}
\begin{figure}
    \centering
    \includegraphics[width=0.97\columnwidth]{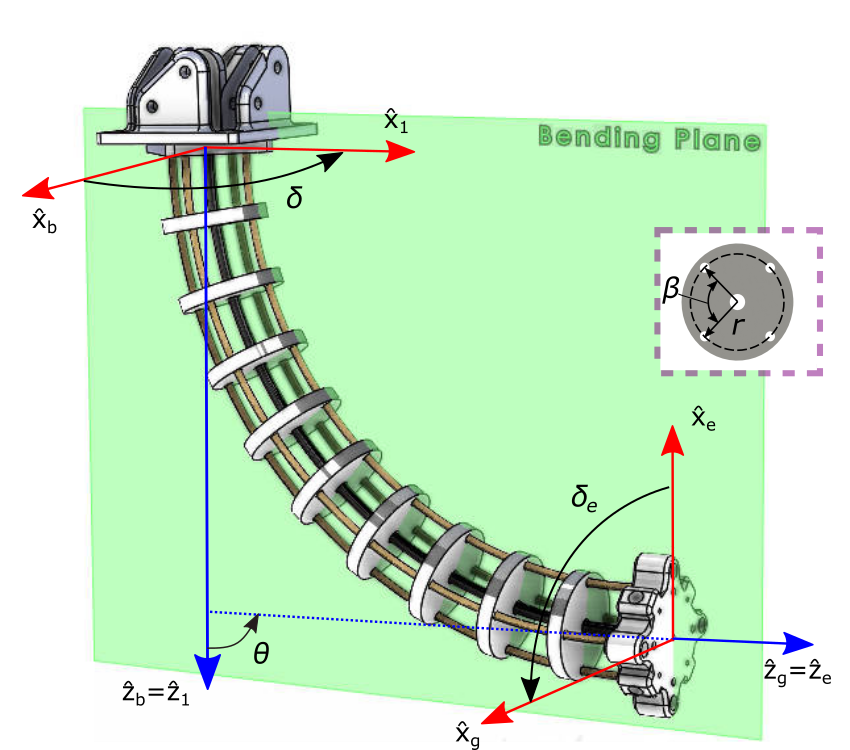}
    \caption{Schematic of the proposed modular continuum arm.(Need to remake)}
    \label{fig:schematic of kinematics}
\end{figure}

We start the kinematic modeling by defining relevant kinematic spaces. A continuum arm kinematics uses three kinematic spaces: the configuration space, the joint space, and the task space. \par
The configuration space is defined as a combination of the in-plane bending angle $\theta$ and the bending plane angle $\delta$, with their vector form as $\psi = [\theta, \delta]^{\mathrm{T}}$. The configuration space is illustrated in Fig.~\ref{fig:schematic of kinematics}. The home (straight) configuration of the proposed continuum arm is \(\psi = [0^\circ, 0^\circ]^{\mathrm{T}}\). \par
Motor positions are denoted by \(q_i\) and they form the joint space. In our case, four motors are used to pull four tendons, \(\mathbf{q} = [ q_1, q_2, q_3, q_4]^{\mathrm{T}} \).\par
Task space indicates the position and orientation of the manipulator's end-effector with respect to the base coordinate. The corresponding homogeneous transformation matrix is as follows:
\begin{equation}
\label{eqn:homo transform}
{ }^{b} \mathbf{T}_{g}={ }^{b} \mathbf{T}_{1}{ }^{1} \mathbf{T}_{e}{ }^{e} \mathbf{T}_{g}
\end{equation}
where\par
\begin{align}
&{ }^{b} \mathbf{T}_{1}=\left[\begin{array}{cc}
\operatorname{RotZ}(\delta) & \mathbf{0} \\
\mathbf{0} & 1
\end{array}\right] \\
&{ }^{1} \mathbf{T}_{e}=\left[\begin{array}{cc}
\operatorname{RotY}(\theta) & { }^{1} \mathbf{p}_{e} \\
\mathbf{0} & 1
\end{array}\right], \quad
{ }^{1} \mathbf{p}_{e}=\frac{L}{\theta}\left[\begin{array}{c}
1-\cos \theta \\[2pt]
0 \label{eqn:kin_p_e}\\[2pt]
\sin \theta
\end{array}\right] \\
&{ }^{e} \mathbf{T}_{g}=\left[\begin{array}{cc}
\operatorname{RotZ}(-\delta) & \mathbf{0} \\
\mathbf{0} & 1
\end{array}\right]
\end{align}
Mappings among these three spaces allows for arm motion control motion. \par
\subsubsection{Configuration to Joint Space}
Inverse kinematics relating the configuration space and joint space is shown as following: 
    \begin{align}
        q_1 &= r\,\cos(\delta)\,\theta \\
        q_2 &= r\,\cos(\delta+\beta)\,\theta \\
        q_3 &= r\,\cos(\delta+2\beta)\,\theta \\
        q_4 &= r\,\cos(\delta+3\beta)\,\theta
    \end{align}
where $ r $ represents the radius of the pitch circle along which the four tendons are circumferentially distributed. We rewrite in vector form for convenience:
\begin{equation}
    \mb{q} = \mb{f}_\mb{q}(\bs{\psi})\label{eqn:kin_joint_space}
\end{equation}

Taking the derivative of (\ref{eqn:kin_joint_space}), the instantaneous inverse kinematics is obtained:
\begin{equation}
    \mathbf{\dot q} = \mb{J}_{\mathbf{q}\psi} \dot{\psi}
\end{equation}
where
\begin{equation}
\mathbf{J}_{\mathbf{q} \psi} =
\begin{bmatrix}
r\, \cos(\delta)                   & -r\, \sin(\delta)\theta \\[6pt]
r\, \cos(\delta+\beta)    & -r\, \sin(\delta+\beta)\theta \\[6pt]
r\, \cos(\delta+2\beta)               & -r\, \sin(\delta+2\beta)\theta \\[6pt]
r\, \cos(\delta+3\beta)    & -r\, \sin(\delta+3\beta)\theta
\end{bmatrix}
\end{equation}
\subsubsection{Configuration to Task Space}

Similarly, the instantaneous direct kinematics can be expressed as:
\begin{equation}
\dot{\mb{x}}=\mb{J}_{\mb{x} \psi} \dot{\boldsymbol{\psi}}, \qquad
\dot{\mb{x}}\triangleq\left[\mb{v}\T,\bs{\omega}\T\right]\T,\quad
\mb{J}_{\mb{x} \psi}\in \realfield{6 \times2}
\end{equation}
where $\dot{\mb{x}}$ represents a twist that consists of a linear velocity $ \mb{v} $ and an angular velocity $ \bs{\omega} $. We further partition and $\mathbf{J}_{\mathbf{x} \psi}$ as:
\begin{equation}
    \mb{J}_{\mb{x}\bs{\psi}} = 
    \left[
        \begin{array}{l}
        \quad\mb{J}_{\mb{v}\bs{\psi}} \quad\\[4pt] \hdashline[2pt/2pt]
        \quad\mb{J}_{\bs{\omega}\bs{\psi}}\quad
        \end{array} 
    \right]
\end{equation}
where $\mb{J}_{\mb{x}\bs{\psi}}$ is the geometric Jacobian matrix that consists of a linear velocity partition and an angular velocity partition. Matrix $\mathbf{J}_{\mathbf{v} \psi}$ can be obtained by differentiating the position ${ }^{b} \mathbf{p}_{e}$. 
Based on (\ref{eqn:kin_p_e}), ${ }^{b} \mathbf{p}_{e}$ may be expressed as:
\begin{equation}
{ }^{b} \mathbf{p}_{e}=
\operatorname{RotZ(\delta)}{ }^{1} \mathbf{p}_{e} = \frac{L}{\theta}\left[\begin{array}{c}
\cos(\delta)\,\left(1-\cos\theta\right) \\[2pt]
\sin(\delta)\,\left(1-\cos\theta\right) \\[2pt]
\sin(\theta)
\end{array}\right]
\end{equation}
Thereby, $\mathbf{J}_{\mathbf{v} \psi}$ is given by
\begin{equation}
\mathbf{J}_{\mathbf{v} \psi}=L\left[\begin{array}{cc}
\cos(\delta)\, \frac{\theta \sin(\theta)+\cos(\theta)-1}{\theta^{2}} & -\frac{\sin(\delta)\,\left(1-\cos(\theta)\right)}{\theta} \\[6pt]
\sin(\delta)\, \frac{\theta \sin(\theta)+\cos(\theta)-1}{\theta^{2}} & \frac{\cos(\delta)\,\left(1-\cos(\theta)\right)}{\theta} \\[6pt]
\frac{\theta \cos(\theta)-\sin(\theta)}{\theta^{2}} & 0
\end{array}\right]
\end{equation}
\par Matrix $\mathbf{J}_{\mathbf{\omega} \psi}$ can be obtained by directly formulating the angular velocity in base coordinate shown as follows
\begin{equation}
{ }^{b} \boldsymbol{\omega}_{g}=\dot{\delta}^{b} \hat{\mathbf{z}}_{b}+{ }^{b} \mathbf{R}_{1}\left(\dot{\theta}^{1} \hat{\mathbf{y}}_{1}+{ }^{1} \mathbf{R}_{e}\left(-\dot{\delta}^{e} \hat{\mathbf{z}}_{e}\right)\right) = \mathbf{J}_{\omega \psi} \dot{\boldsymbol{\psi}}
\label{eqn:J_omega_psi_derivation}
\end{equation}
where $^{b}\hat{\mathbf{z}}_{b} = [0, 0, 1]^{\mathrm{T}}$, $^{1}\hat{\mathbf{y}}_{1} = [0, 1, 0]^{\mathrm{T}}$, $^{e}\hat{\mathbf{z}}_{e} = [0, 0, 1]^{\mathrm{T}}$ in their respective coordinates. ${ }^{b}\mathbf{R}_{1}$ and ${ }^{1} \mathbf{R}_{e}$ can be obtained from (\ref{eqn:kin_p_e}).
By rearranging (\ref{eqn:J_omega_psi_derivation}) in a vector format, matrix $\mathbf{J}_{\omega \psi}$ can be given by
\begin{equation}
\mathbf{J}_{\omega \psi}=\left[\begin{array}{cc}
-\sin(\delta)& -\cos(\delta)\,\sin(\theta)   \\
\cos(\delta) & -\sin(\delta)\,\sin(\theta)  \\
0& 1-\cos(\theta) 
\end{array}\right]
\end{equation}
\subsection{Statics}
We use the virtual work analysis to derive the statics in this section. We take the potential elasticity energy into account. When the continuum arm bends, the central NiTi backbone stores the potential elastic energy, given by:

\begin{equation}
\mb{E}=\int_{L} \frac{E I}{2}\left(\frac{d \theta}{d s}\right)^{2} d s=\theta^{2}\frac{E_{p} I_{p}}{2 L}
\label{eqn:elastic_energy}
\end{equation}
\par Following the derivation method in \cite{simaan2005snake,Goldman2014}, the statics model of the continuum arm can be described as:
\begin{equation}
\label{eqn:static equilibrium}
(\mb{J}_{\mb{q} \psi})\T \;\bs{\tau}\; +\; (\mb{J}_{\mb{x} \psi})\T \;\mb{w}_{\text{ext}}\; =\; \nabla \mb{E} 
\end{equation}
where $\bs{\tau}$ represents the actuation forces for the tendon pulling, $\mb{w}_\text{ext}$ is the 6-dimensional external wrench applied to the end-disk of the continuum arm, $\nabla \mb{E}$ describes the gradient of the elastic energy with respect to the configuration space change.\par 
By differentiating (\ref{eqn:elastic_energy}), $\nabla \mb{E}$ takes the form:
\begin{equation}
\begin{aligned}
&\nabla \mb{E}
&=\left[\begin{array}{c}
\frac{\theta E_{p} I_{p}}{L}\\[4pt]
0
\end{array}\right]
\end{aligned}
\end{equation}
\input{content/stiffness_modeling}

%% file: content/stiffness_modeling.tex
\subsection{Stiffness}
Stiffness modeling of a continuum manipulator is essential for the case of using it in aerial manipulation. Without considering the reaction moment and orientational displacement, we limit the scope of the stiffness modeling problem to incremental force and positional displacement. \par
\begin{figure}[!h]
	\centering
	\includegraphics[width=0.75\columnwidth]{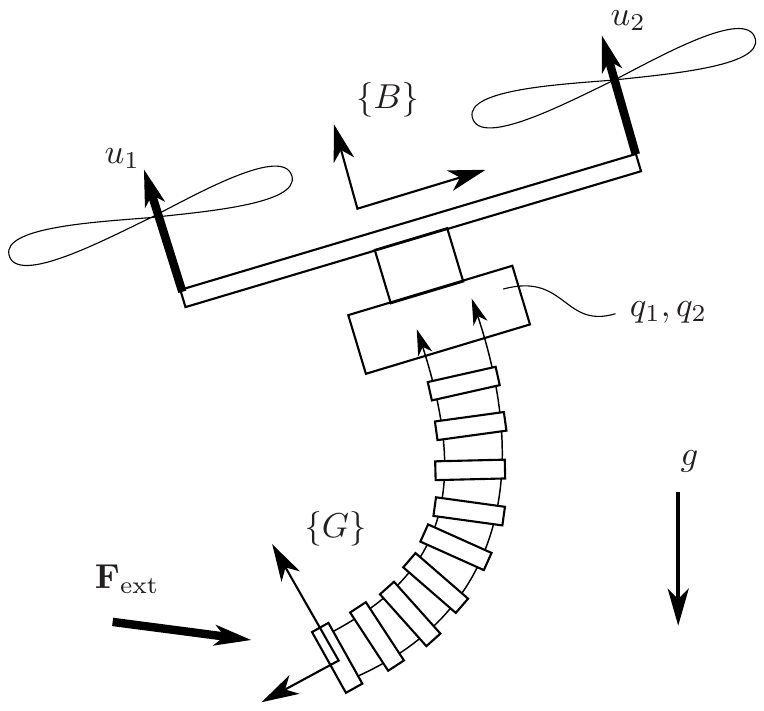}
	\caption{Stiffness modeling of a continuum manipulator for a drone}
	\label{fig:stiffness_modeling_aerial}
\end{figure}\par
As illustrated in Fig.~\ref{fig:stiffness_modeling_aerial}, let $ \mb{F}_\text{ext} $ and $ {}^b\mb{p}_g $ both be $ (3\times1) $ vectors, representing an external force and the gripper position expressed in the manipulator base frame, respectively. Thereby, the ultimate goal of stiffness modeling is to capture a $ 2^{\text{nd}} $-order gradient tensor, $ \mb{K}_X $, defined as:
\begin{equation}
	\mb{K}_X \triangleq \dfrac{\partial  (\mb{F}_{\text{ext}})}{\partial  ({}^b\mb{p}_g)}, \qquad \mb{K}_X\in\realfield{3\times3}
\end{equation}\par
The stiffness matrix $ \mb{K}_X $ (the $ 2^{\text{nd}} $-order gradient tensor) defines the linearized local relations between small force change $ \rmd\mb{F} $ and small deformation $ \rmd \mb{p} $. However, the derivation of $ \mb{K}_X $ is nontrivial for continuum robots due to the highly nonlinear mappings among the three kinematic spaces as discussed earlier. \par
A stiffness matrix in the configuration space can be found only assuming that the external force is very small \cite{Goldman2014}, and it is thereby defined as:
\begin{equation}
	\mb{K}_\psi \triangleq \frac{\partial \mb{F}^{*}}{\partial \bs{\psi}}, \qquad \mb{K}_\psi\in\realfield{2\times2}
	\label{eqn:k_psi}
\end{equation}
where $ \mb{F}^{*} $ denotes the generalized force in configuration space when the external wrench (including force) is very small. By considering the static equilibrium state from (\ref{eqn:static equilibrium}), we can express such a generalized force applied to the continuum arm in configuration space, as:
\begin{equation}
 \mb{F}^{*} =\nabla \mb{E}-(\mb{J}_{\mb{q} \psi})\T \; \bs{\tau}=(\mb{J}_{\mb{x} \psi})\T\; \mb{w}_{\text{ext}}\approx\mb{0},\qquad 
 \mb{F}^{*} \in \realfield{2\times1}
\end{equation}\par 
Following \cite{Goldman2014}, one could find the expression of the aforementioned configuration space stiffness matrix, $ \mb{K}_\psi $, as:
\begin{equation}
\mb{K}_{\bs{\psi}}=\mb{H}_{\bs{\psi}}-\left[\frac{\partial}{\partial \bs{\psi}}\left(\mb{J}_{\mb{q} \psi}\right)\T\right] \bs{\tau}-(\mb{J}_{\mb{q} \psi})\T \;\mb{K}_{\mb{q}} \; (\mb{J}_{\mb{q} \psi})
\end{equation}
where 
\begin{align}
	& \mb{H}_{\bs{\psi}} = \left[\begin{array}{cc}
\dfrac{\partial^{2} \mb{E}}{\partial \theta^{2}} &
\dfrac{\partial^{2} \mb{E}}{\partial \theta \partial \delta}\\[8pt]
\dfrac{\partial^{2} \mb{E}}{\partial \delta \partial \theta}&
\dfrac{\partial^{2} \mb{E}}{\partial \delta^{2}}
\end{array}\right]= 
\left[\begin{array}{cc}
\dfrac{ E_{p} I_{p}}{L} &0\\[8pt]
0& 0
\end{array}\right]
\\
& \mb{K}_q= \operatorname{diag}^{4}\left(\left[\frac{E_{T} A}{L}, \ldots, \frac{E_{T} A}{L}\right]\right)
\end{align}

Note that in the above equation, $\left[\frac{\partial}{\partial \bs{\psi}}\left(\mb{J}_{\mb{q} \psi}\right)\T\right]$ is a $3^{rd}$-order tensor (or a ``3D matrix''), and the multiplication is facilitated as:
\begin{equation}
    \left[\dfrac{\partial}{\partial \bs{\psi}}\left(\mb{J}_{\mb{q} \psi}\right)\T\right] \bs{\tau}\quad =\quad
    \left[
        \begin{array}{c;{2pt/2pt}c}
	         \dfrac{\partial (\mb{J}_{\mb{q} \psi})\T}{\partial {\theta}} \; \bs{\tau} \;\;  & \;\;  
	         \dfrac{\partial (\mb{J}_{\mb{q} \psi})\T}{\partial {\delta}} \; \bs{\tau} \;\;
	        \end{array}
    \right]
\end{equation}

We can find the expression of $ \mb{K}_X $ if we keep using the small force assumption. Let us express the small force using the generalized force, as:

\begin{equation}
    \label{eqn: external force}
 	\mb{F}^{*}=(\mb{J}_{\mb{v} \bs{\psi}})\T \; \mb{F}_{\text{ext}} \quad \rightarrow \quad \mb{F}_\text{ext}=\left(
	\mb{J}_{\mb{v} \psi}\T
	\right)^{\dagger}
	\mb{F}^{*}
\end{equation}

\par Thus, the stiffness matrix under task space is given by
\begin{equation}
	\begin{array}{rl}
		\mb{K}_{X} & =\frac{\partial \mb{F}_{\text{ext}}}{\partial ({}^b\mb{p}_g)}
		= \frac{\partial \left(
			\left(
			\mb{J}_{\mb{v} \psi}\T
			\right)^{\dagger}
			\mb{F}^{*}\right)}{\partial \bs{\psi}}\frac{ \partial \bs{\psi}}{\partial ({}^b\mb{p}_g)} \\[12pt]
		&= \left[
		\frac{\partial}{\partial \bs{\psi}}\left(\mb{J}_{\mb{v} \psi}\T\right)^{\dagger}
		\right]
		\mb{F}^{*}\mb{J}_{\mb{v} \psi}^{\dagger}+\left(\mb{J}_{\mb{v} \psi}\T\right)^{\dagger}\mb{K}_{\mb{\psi}}\mb{J}_{\mb{v} \psi}^{\dagger}
	\end{array}
\end{equation}
where $\left[\frac{\partial}{\partial \bs{\psi}}\left(\mb{J}_{\mb{v}\psi}\T\right)^{\dagger}\right]$ is a $3^{rd}$-order tensor (or a ``3D matrix'').

%% file: content/imu_force_estimation.tex
\section{IMU-based Force Estimation}
Inertia Measurement Unit (IMU) sensors have become a ubiquitous sensory feedback source for robotics, thanks to the fact that they are low-cost, compact, and easy to deploy. In this work, we wish to explore its usage for estimating force during manipulation and physical interactions.  

\subsection{IMU Sensor Integration}\label{sec:imu_sensor_integration}
\begin{figure}[!h]
	\centering
	\includegraphics[width=1\columnwidth]{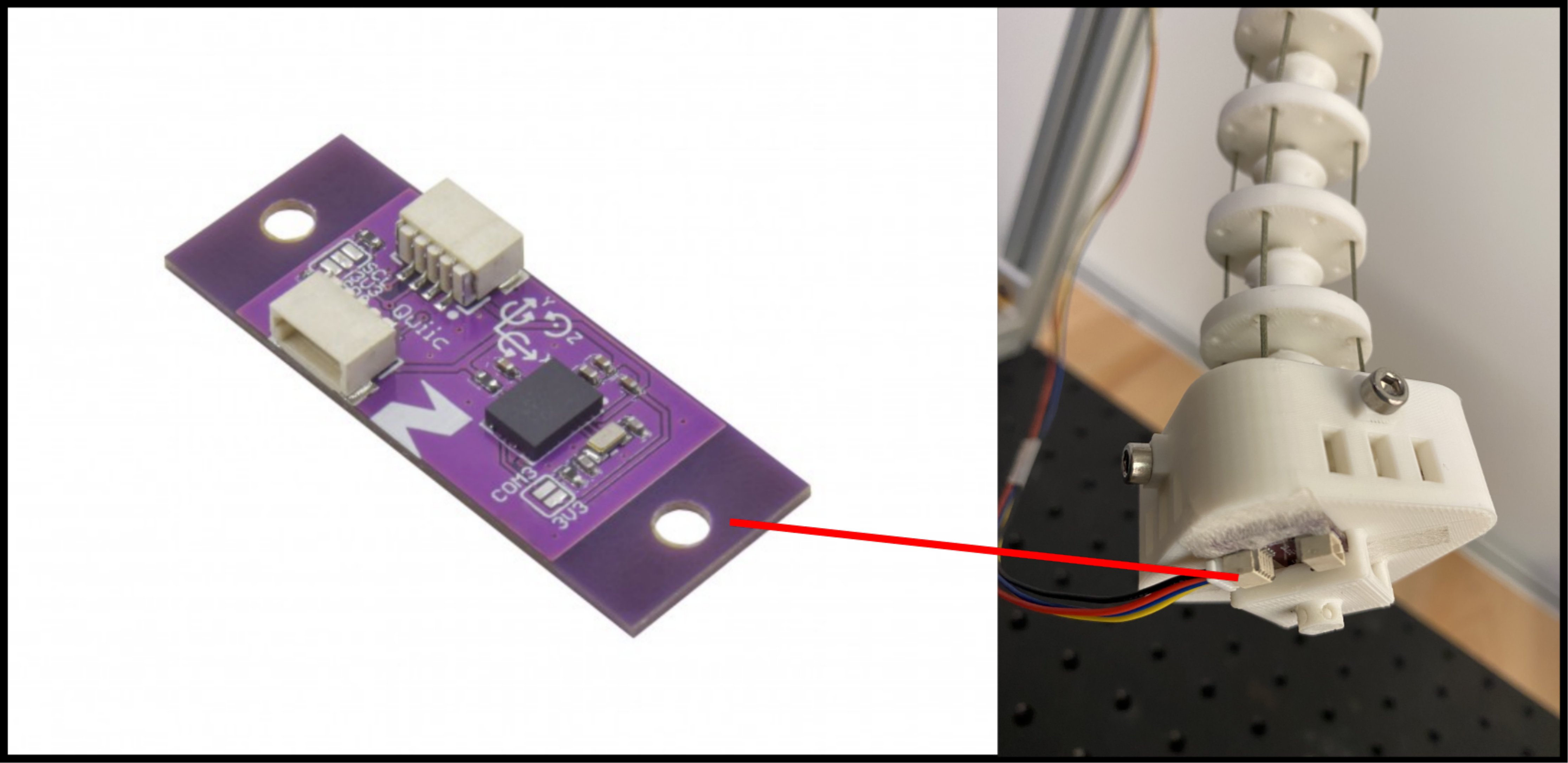}
	\caption{Integration of an IMU sensor to the end-disk of the continuum manipulator}
	\label{fig:imu_sensor_integration}
\end{figure}

A low-cost IMU sensor (ZIO 9DOF IMU BNO080) was integrated to the end-disk of the continuum manipulator, and it provided orientation measurements in real-time. As illustrated in Fig.~\ref{fig:imu_sensor_integration}, the IMU sensor was rigidly attached to the end-disk of the arm. 
\par
Then, the available IMU measured orientations can be used for the proposed force estimation algorithm, which is discussed as follows.

\subsection{Force Estimation Algorithm}\label{sec:force_est_algorithm}

The overall workflow of the framework is: we predict the external forces applied to the arm tip using the stiffness matrix and the deformation measured from an IMU sensor. \par
%

Let us assume that the real-time orientation measurements can be obtained with respect to the base frame of the continuum robot, denoted as:
\begin{equation}
    \overline{\mb{R}} = \begin{bmatrix}
    r_{11} & r_{12} & r_{13} \\
    r_{21} & r_{22} & r_{23} \\
    r_{31} & r_{32} & r_{33}
    \end{bmatrix}
\end{equation}
This orientation measurement can be obtained via rotating the IMU-provided orientations by a known offset between the base of the robot and the world inertia frame.\par

With the real-time orientation measurements, it is possible to project those 3D orientations onto the bending plane. Thereby, we could obtain the deformation in configuration space, as:
\begin{align}
    & \overline{\theta} = \text{acos}(r_{33}), \quad \overline{\delta} = \text{atan2}(r_{23},\;r_{13}) \label{eqn:acos_atan2_theta_delta}\\
    & \Delta\overline{ \bs{\psi}} \triangleq \left[\Delta\overline{ \theta}, \;\Delta\overline{ \delta}\right]\T
    \label{eqn:deform_meas_psi}
\end{align}
We had recognized the limitation and inaccuracy of obtaining the configuration space deformation using the direct projection calculation as in (\ref{eqn:acos_atan2_theta_delta}) and (\ref{eqn:deform_meas_psi}). We plan to investigate more elegant observer-based approaches in the future to provide more robust estimates of $\Delta\overline{ \bs{\psi}}$. \par 
With the measured configuration space deformation $\Delta\overline{ \bs{\psi}}$ and the computation of the stiffness matrix $\mb{K}_\psi$ from (\ref{eqn:k_psi}), we can estimate the generalized forces in configuration space:
\begin{equation}
    \hat{\mb{F}}^{*} = \mb{K}_\psi \; \Delta\overline{ \bs{\psi}}
\end{equation}
Then, following (\ref{eqn: external force}), we chose to use the geometric Jacobians to transmit the configuration space generalized forces to the task space end-effector forces:
\begin{equation}
    \hat{\mb{F}}_\text{ext} = \left(
	\mb{J}_{\mb{v} \psi}^{\mathrm{T}}
	\right)^{\dagger}\; \hat{\mb{F}}^{*}
\end{equation}\par
\begin{figure}[!h]
    \centering
    \includegraphics[width=1\textwidth]{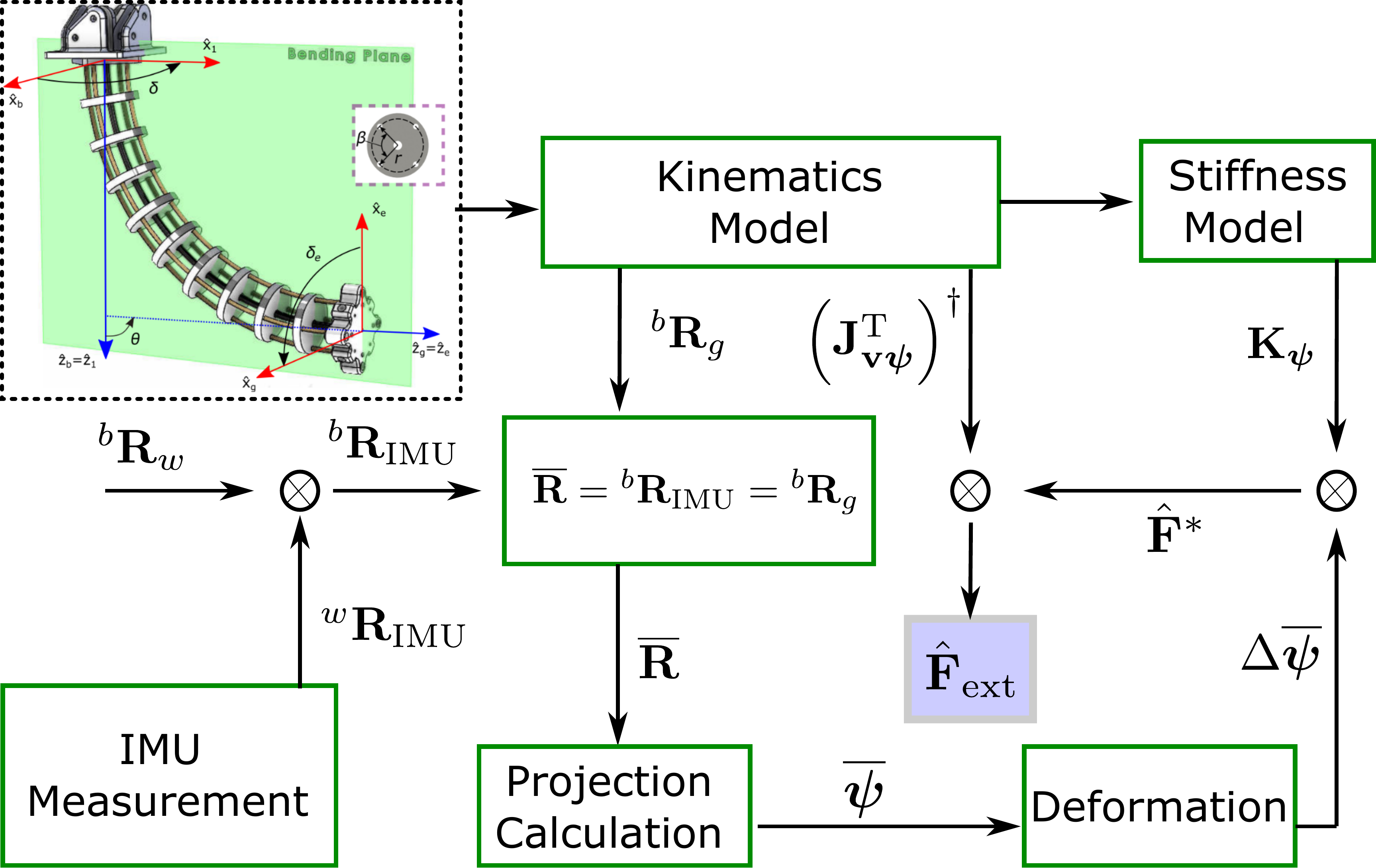}
    \vspace{-1mm}
    \caption{Computation workflow of the IMU-based force estimation framework}
    \label{fig:framework_imu_force_estimation}
\end{figure}

In Fig.~\ref{fig:framework_imu_force_estimation}, we illustrate the computation workflow of the proposed IMU-based force estimation.

%% file: content/experiments.tex
\section{Preliminary Experimental Validation}
\label{section:experiments}
In this section, we report preliminary results on three experiments:
\begin{enumerate}[label=Exp-\arabic*]
    \item \textit{Benchmark Perching}: we demonstrate the feasibility of perching and the benefit of softening contact impact 
    \label{idx:exp_perching}
    \item \textit{Arm Stiffness}: we report quantitative stiffness (compliance) metrics for the particular arm design chosen in this work \label{idx:exp_arm_stiffness}
    \item \textit{IMU-based Force Estimation}: we demonstrate the feasibility of estimating force using the IMU sensor integrated \label{idx:exp_imu_force}
\end{enumerate}
\input{content/exp_perching}

\input{content/exp_arm_stiffness}
\input{content/exp_imu_force_est}

%% file: content/exp_perching.tex
\subsection{Benchmark Perching Experiment}

Mechanical compliance of continuum arms enables safe physical interaction with the surroundings. 
When the continuum arm is in motion and starts colliding, this safe feature is reflected in the slope of the measured force signal: a hard hit behaves as a step function with a steep slope while a soft hit behaves as a gentle slope.  
%
This experiment aims to qualitatively demonstrate the safe (soft hit) behavior by revealing the magnitude of force disturbance exerted to the aerial continuum arm's carrier during contact initiation process. \par
\begin{figure}[!h]
    \vspace{5mm}
    \centering
    \includegraphics[width=0.8\textwidth]{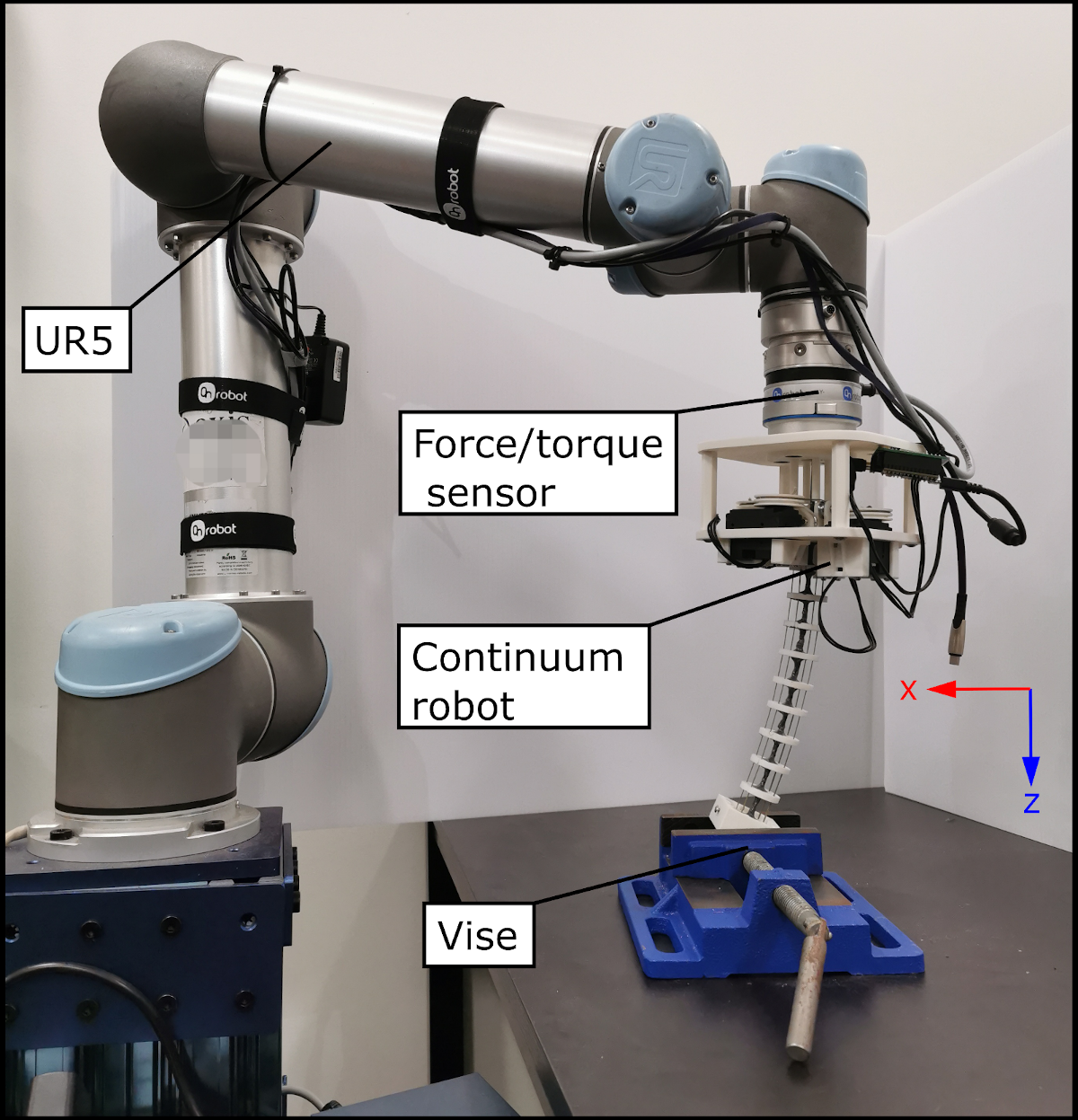}
    \caption{Benchmark Perching Experiment (\ref{idx:exp_perching}) setup: the in-door aerial continuum arm testing platform consists of a UR5 robotic arm, a F/T sensor, and the aerial continuum arm system.}
    \label{fig:benchmark exp setup}
\end{figure}\par
Instead of testing the initial arm prototype on a flying UAV, we developed a mock-up indoor test environment (as depicted in Fig.~\ref{fig:benchmark exp setup}). We used a collaborative robot arm - Universal Robot UR5 - to simulate the UAV flight motion. A 6-axis force/torque (F/T) sensor was mounted on UR5's end-effector to provide the real-time measurements. \par

\begin{figure*}[!h]
\setlength\Myfigwd{0.72\columnwidth}
\floatbox[{\capbeside\thisfloatsetup{
capbesideposition={right,center},
capbesidewidth=\dimexpr\linewidth-\Myfigwd-3em\relax}}]{figure}[\FBwidth]
{\caption{Benchmark Perching Experiment (\ref{idx:exp_perching}) results: the reaction force and torque measured by a F/T sensor while UR5 end-effector travels in (a) x-direction, and (b) z-direction. 
The red dash line on each plot denotes robot trajectory and refers to the y-axis on the right.
}
\label{fig:benchmark results}}
{\includegraphics[width=\Myfigwd]{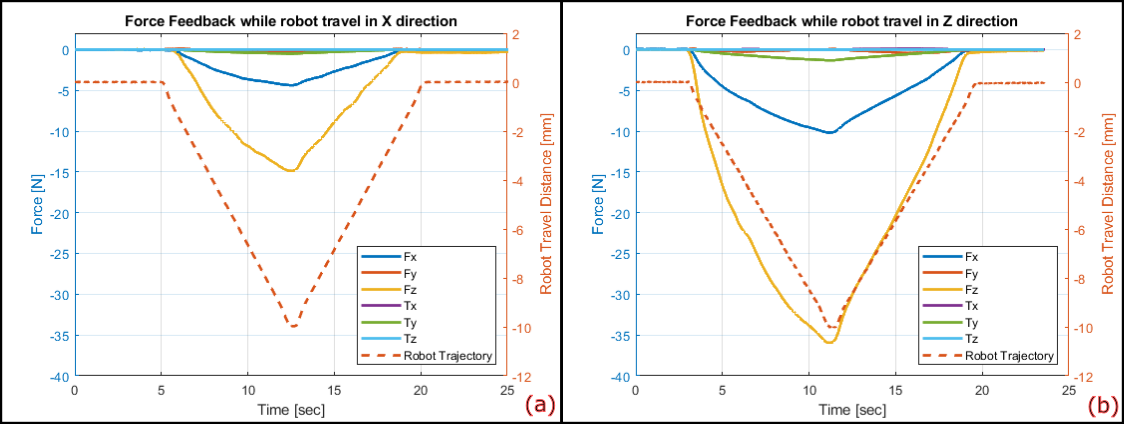}}
\end{figure*}

During the experiment, the continuum arm bent at an angle of 30 degrees, and its tip was secured at a heavy-duty bench vise on a table. UR5 was programmed to travel 10 mm in its x-direction and then returned to its starting position with the same speed. The corresponding force measurements at the F/T sensor were recorded for analysis. The same experiment procedure was repeated with proceeding 10 mm in the z-direction and returned. \par
%
The benchmark perching experiment results are shown in Fig.~\ref{fig:benchmark results}: (a) and (b) with motion in x and z directions, respectively. The negative force and torque values indicate that they were applied in the opposite direction and were referred to the left y-axis of the plot. The red dash line indicates the robot trajectory and refers to the y-axis on the right. As expected, the force exerted to the upper structure of the continuum arm system (in this case, the F/T sensor and UR5 end-effector) gradually increased and decreased, achieving a gentle slope behavior. Results indicate that, when the arm is in contact with environment, the compliance of the continuum arm helps soften the force disturbance and avoids a high impact to the aerial arm carrier over a short time period. \par

%% file: content/exp_arm_stiffness.tex
\subsection{Arm Stiffness Experiment}

The proposed continuum arm design structure is low-cost and highly customizable: changing thickness of the central NiTinol backbones, (optionally) increasing number of surrounding NiTinol backbones, changing radius of the pitch circle of parallel NiTinol backgones, and more. The modeling work presented earlier can facilitate the calculation of structural stiffness given selected customizable parameters, which serves as a guideline for generating designs for different tasks. \par 
In this section, we present the experimental apparatus to validate the stiffness of a chosen and prototyped design. The experimental validation can help correct prior assumptions on parameter values. We choose stiffness as the key metric to be evaluated experimentally.
%
Stiffness is the correspondence of external force and arm displacement. 

\begin{figure}[!h]
    \centering
    \includegraphics[width=1\textwidth]{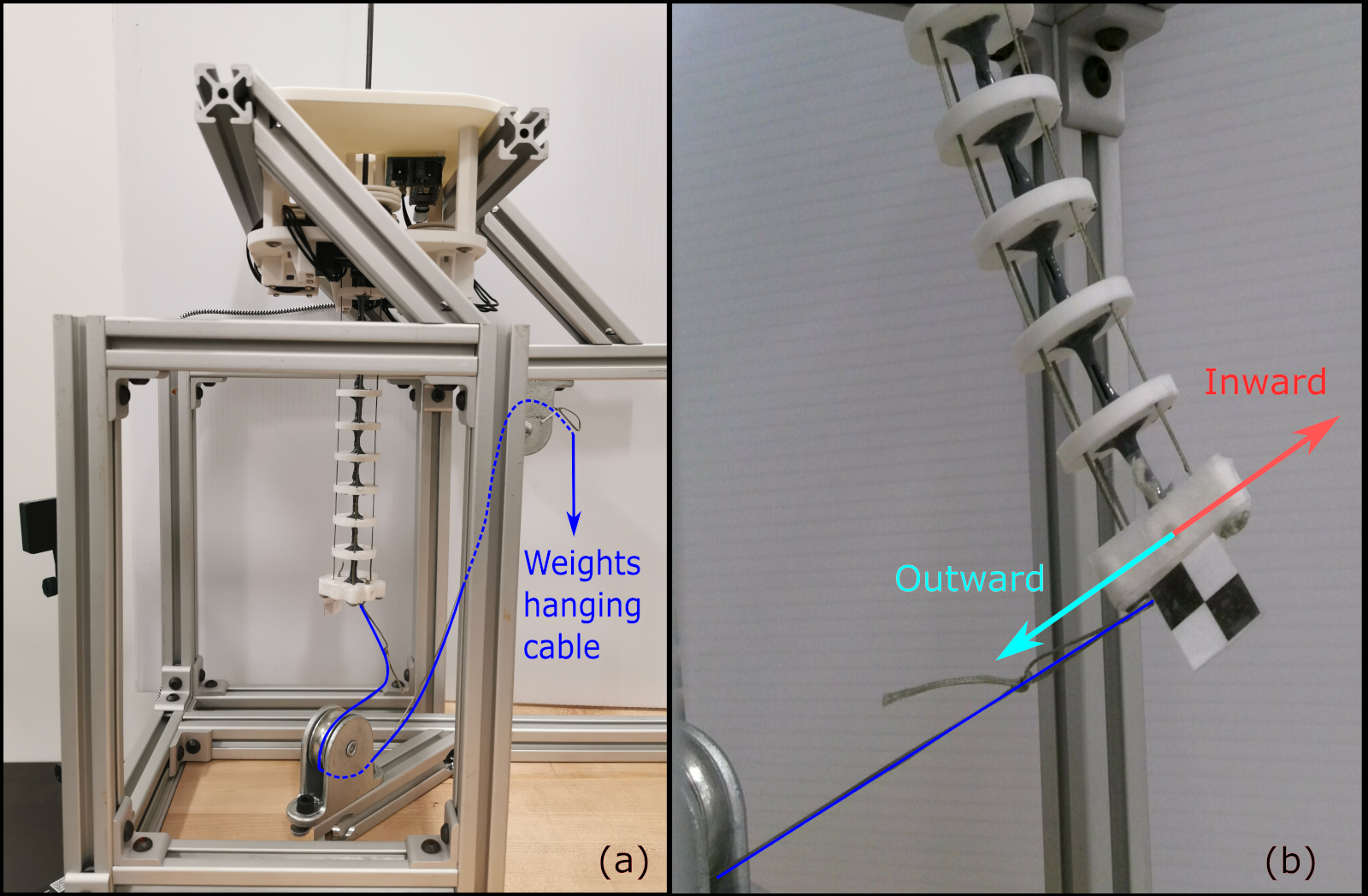}
    \caption{Arm Stiffness Experiment (\ref{idx:exp_arm_stiffness}) setup: (a) is the test setup of aerial continuum arm stiffness evaluation under a known load. The top plate of the arm system is fixed with the testing structure, and a known load is applied at the tip of the arm. (b) illustrates of inward and outward radial loading.}
    \label{fig:stiffness test setup}
\end{figure}

The stiffness experimental setup is shown in Fig.~\ref{fig:stiffness test setup}(a). Pulleys were adjustable along the rail to ensure that the external load was applied perpendicularly to the tip of the arm. The adapter plate was fastened down at the top of the structure to ensure that the base stayed stationary during the test. The blue line in Fig.~\ref{fig:stiffness test setup}(a) indicates the weights hanging cable and its route through pulleys in the inward radial loading condition (within the bending plane, the load is applied with the bending angle). This setup is also designed to flip the loading direction for outward loading experiments (within the bending plane, the load is applied against the bending angle). Figure~\ref{fig:stiffness test setup}(b) indicates these two loading conditions. \par

\begin{figure*}[!h]
\setlength\Myfigwd{0.75\columnwidth}
\floatbox[{\capbeside\thisfloatsetup{
capbesideposition={right,center},
capbesidewidth=\dimexpr\linewidth-\Myfigwd-3em\relax}}]{figure}[\FBwidth]
{\caption{Arm Stiffness Experiment (\ref{idx:exp_arm_stiffness}) results: (a) shows the results of inward radial loading; (b) shows the results of outward radial loading.}
\label{fig:stiffness results}}
{\includegraphics[width=\Myfigwd]{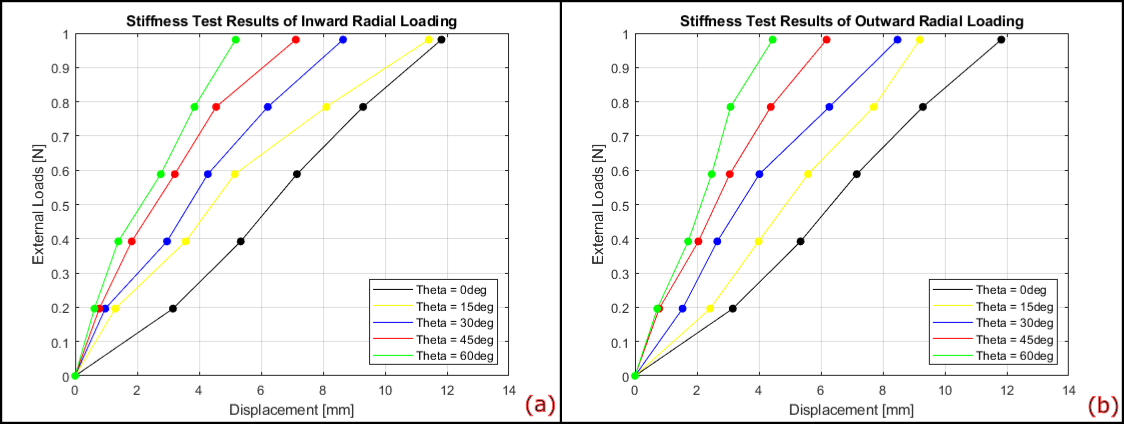}}
\end{figure*}

The stiffness tests were repeated at five different arm configurations on each loading condition, and results are shown in Fig.~\ref{fig:stiffness results}, where (a) indicates results of inward loadings and (b) for outward loadings. An external load of up to approximately 1N was applied perpendicular to the bending direction, with five loading cycles and 20 gram increments at each time. The arm straight configuration (denoted in black lines in the plots) had the lowest stiffness. Generally as the bending angle increased, the arm stiffened in both loading scenarios, which is consistent with what's expected from the stiffness modeling formulation presented earlier.\par

%% file: content/exp_imu_force_est.tex
\subsection{IMU-based Force Estimation Experiment}\label{sec:force est exp}
In this experiment, we demonstrated the real-time estimation of an external force applied to the end-disk using the method presented in Section~\ref{sec:force_est_algorithm}.\par

\begin{figure}
    \vspace{5mm}
    \centering
    \includegraphics[width=1\textwidth]{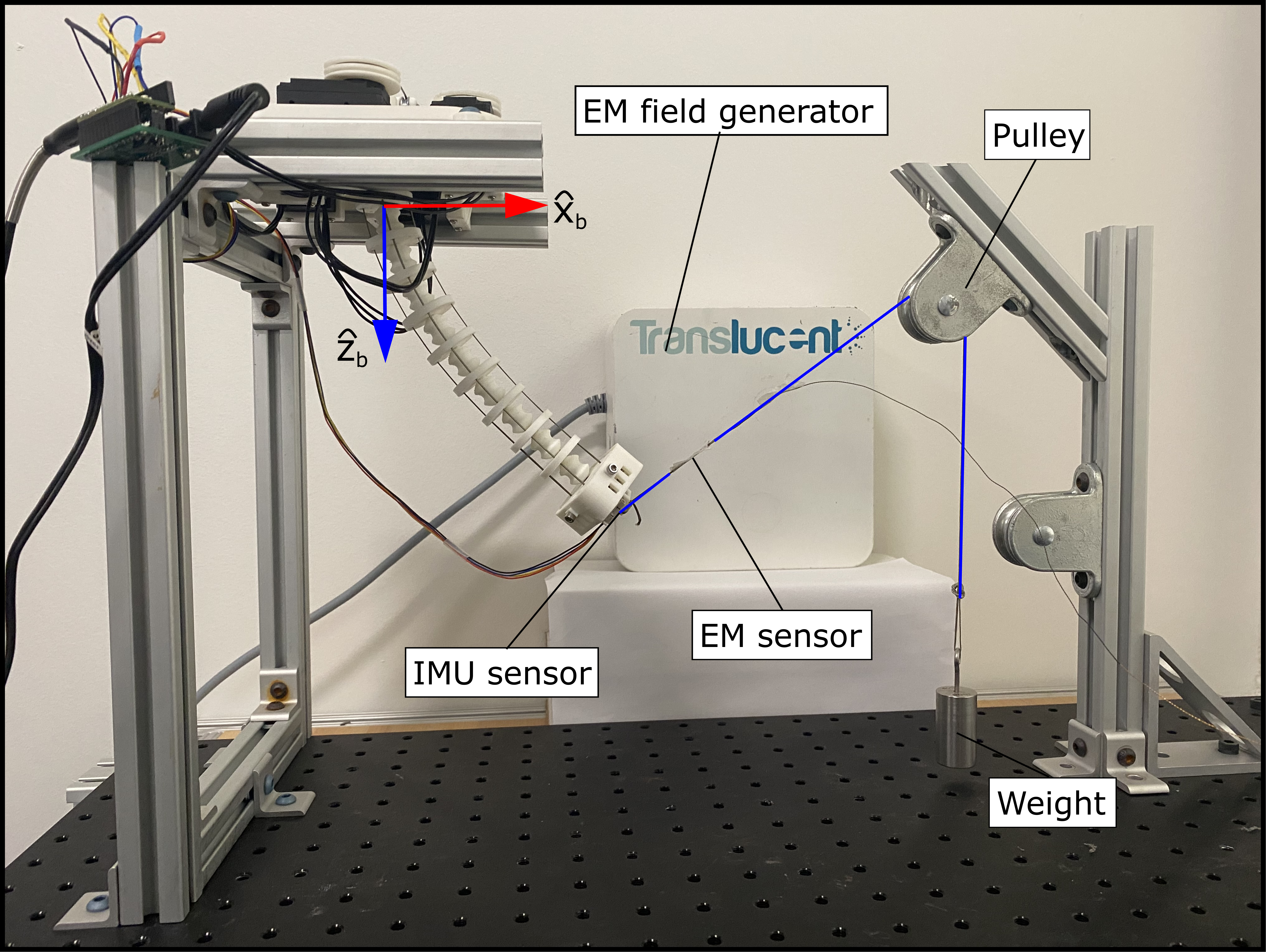}
    \caption{IMU-based Force Estimation Experiment (\ref{idx:exp_imu_force}) setup: an arm prototype with spacers integrated with IMU sensor at the end-disk is fixed on an aluminum frame. a pulley system with weights provides the tip force. And an Electromagnetic tracking system is adopted to measure the direction of the applied force.}
    \label{fig:IMU force exp setup}
\end{figure}

The experimental setup is shown in Fig.~\ref{fig:IMU force exp setup}. To minimize the variation of manufacturing, we adopted the arm prototype with spacers to evenly distribute disks along the central backbone. The IMU sensor used is described in Section~\ref{sec:imu_sensor_integration}. The force applied to end-disk was provided by hanging weights through a pulley system, and the force was always parallel to the $\uvec{x}_b$-$\uvec{z}_b$ plane of Frame ${\{B\}}$. The external loads (hanging weights) varied from 20g to 100g. The direction of the applied force was measured by an Electromagnetic (EM) tracking system (NDI Aurora V2). The EM sensor was small and light enough to be attached to the loading cable of hanging the weight with negligible impact to the force exerted direction. \par
The continuum robot was controlled to articulate in different configurations. In total, 4 configurations were considered for inward and outward loading cases. The basic testing workflow for each configuration is as follows: (1) obtain applied force vector by combining the information from EM sensor measurement and the hanging weight; (2) turn off EM tracking system and start IMU sensor reading in real-time; (3) calculate the change of configuration space $\Delta{\bs{\psi}}$ under different applied forces; (4) compute the external force in configuration space $\mb{F}^{*}$ based on the derived stiffness model; (5) compute the external force in task space $\mb{F}_\text{ext}$ using equation (\ref{eqn: external force}); (6) repeat steps 2-5 three times and get averaged estimated force.\par
\begin{figure*}[!h]
\setlength\Myfigwd{0.71\columnwidth}
\floatbox[{\capbeside\thisfloatsetup{
capbesideposition={right,center},
capbesidewidth=\dimexpr\linewidth-\Myfigwd-3em\relax}}]{figure}[\FBwidth]
{\caption{IMU-based Force Estimation Experiment (\ref{idx:exp_imu_force}) results under different configuration space conditions: (a) red: $\theta = 45 ^{\circ}, \delta = 0 ^{\circ}$, blue: $\theta = 45 ^{\circ}, \delta = 180 ^{\circ}$  (b) red: $\theta = 30 ^{\circ}, \delta = 0 ^{\circ}$, blue:$\theta = 30 ^{\circ}, \delta = 180 ^{\circ}$}
\label{fig:force_sensing results}}
{\includegraphics[width=0.99\Myfigwd]{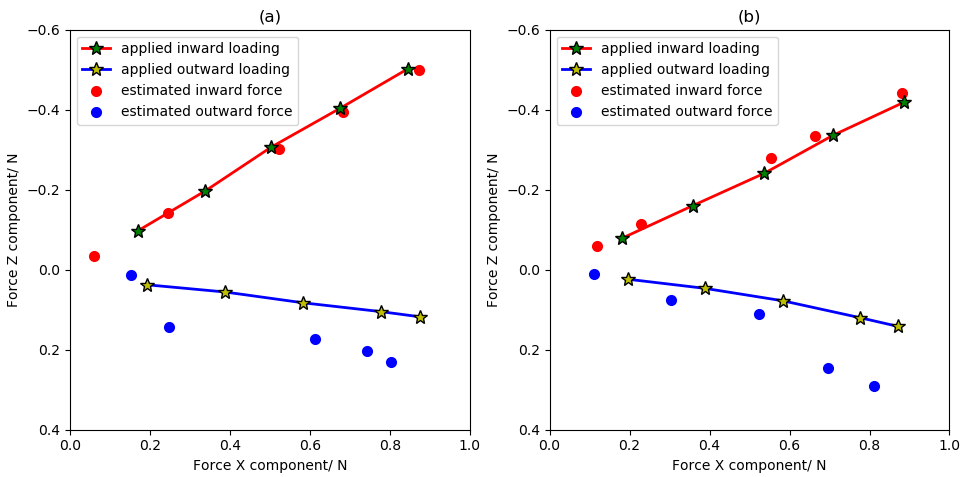}}
\end{figure*}

The results are detailed in Table~\ref{tab:IMU results} and illustrated in Fig.~\ref{fig:force_sensing results}. Table~\ref{tab:IMU results} reports the force estimates, the actual applied force projected in $\uvec{x}_b$ and $\uvec{z}_b$, and the associated errors. In the inward loading cases, the average error is 0.039N with a standard deviation 0.036N. In the outward loading cases, the average error is 0.072N with a standard deviation 0.039N. Figure~\ref{fig:force_sensing results} shows plots of the results of force estimation. \par
The force estimation method has a better performance in the inward loading cases than in the outward loading cases, which we attempt to explain as follows. Let us begin with an underlying simplification in this preliminary work for the force estimation problem. When computing the stiffness matrix in this work, as a simplified starting point, we have ignored the \textit{Active Stiffness} portion that is proportional to the tendon tensions. In other words, we assumed that the tendon tensions are zeros in calculating the stiffness matrix $\mb{K}_{\psi}$:
\begin{equation}
\mb{K}_{\bs{\psi}}=\mb{H}_{\bs{\psi}}-
\underbrace{
\left[\frac{\partial}{\partial \bs{\psi}}\left(\mb{J}_{\mb{q} \psi}\right)\T\right] \big(\; \cancelto{\mb{0}}{\bs{\tau}\;\;}\quad \big)}_{\text{Active Stiffness}=\mb{0}}-(\mb{J}_{\mb{q} \psi})\T \;\mb{K}_{\mb{q}} \; (\mb{J}_{\mb{q} \psi})
\end{equation}\par
As a result, the computed stiffness matrix is more accurate for the ``passive'' cases, in which tendons are slack. In the outward loading cases, the external loading force will increase the tendon tension, which will increase the effect of the active stiffness portion, making the estimation less accurate. Currently, the limitation of not considering the active stiffness part is caused by (i) lack of access to reliable tendon tension measurements as we use low-cost servo motors and (ii) lack of investigations on model-based prediction calculations for tendon tension estimates. We plan to develop methods in the future to access such tendon tension information, in order to compute the active stiffness part.

\vspace{-5mm}

\input{content/tab_force_sensing}

%% file: content/tab_force_sensing.tex
\begin{table*}[h]
\footnotesize
\centering
\captionsetup{justification=centering}
\caption{IMU-based Force Estimation Experiment (\ref{idx:exp_imu_force}) results. (All units are Newton)}
\begin{tabular}{llllllllllll}
\thickhline
\multicolumn{6}{|c|}{Inward loading case I}                                                                                                                                                          & \multicolumn{6}{c|}{Outward loading case I}                                                                                                                                                          \\
\multicolumn{6}{|c|}{$\theta = 45^{\circ}, \delta = 0^{\circ}$}                                                                                                                                                     & \multicolumn{6}{c|}{$\theta = 45^{\circ}, \delta = 180^{\circ}$}                                                                                                                                                    \\ \hline
\multicolumn{2}{|c|}{Applied Force}                             & \multicolumn{2}{c|}{Estimated Force}                               & \multicolumn{2}{c|}{Error}                                      & \multicolumn{2}{c|}{Applied Force}                              & \multicolumn{2}{c|}{Estimated Force}                               & \multicolumn{2}{c|}{Error}                                      \\ 
\multicolumn{1}{|c|}{\textbf{x}}        & \multicolumn{1}{c|}{\textbf{z}}         & \multicolumn{1}{c|}{\textbf{x}}        & \multicolumn{1}{c|}{\textbf{z}}         & \multicolumn{1}{c|}{\textbf{x}}        & \multicolumn{1}{c|}{\textbf{z}}        & \multicolumn{1}{c|}{\textbf{x}}        & \multicolumn{1}{c|}{\textbf{z}}        & \multicolumn{1}{c|}{\textbf{x}}         & \multicolumn{1}{c|}{\textbf{z}}         & \multicolumn{1}{c|}{\textbf{x}}        & \multicolumn{1}{c|}{\textbf{z}}         \\ \hline
\cellcolor[HTML]{C0C0C0}0.170  & -0.097                         & \cellcolor[HTML]{C0C0C0}0.060  & -0.035                         & \cellcolor[HTML]{C0C0C0}0.110  & -0.062                         & \cellcolor[HTML]{C0C0C0}0.193  & 0.038                          & \cellcolor[HTML]{C0C0C0}0.152  & 0.013                          & \cellcolor[HTML]{C0C0C0}0.041  & 0.025                          \\
\cellcolor[HTML]{C0C0C0}0.339  & -0.197                         & \cellcolor[HTML]{C0C0C0}0.245  & -0.141                         & \cellcolor[HTML]{C0C0C0}0.094  & -0.056                         & \cellcolor[HTML]{C0C0C0}0.388  & 0.056                          & \cellcolor[HTML]{C0C0C0}0.247  & 0.144                          & \cellcolor[HTML]{C0C0C0}0.141  & -0.088                         \\
\cellcolor[HTML]{C0C0C0}0.503  & -0.306                         & \cellcolor[HTML]{C0C0C0}0.522  & -0.302                         & \cellcolor[HTML]{C0C0C0}-0.019 & -0.004                         & \cellcolor[HTML]{C0C0C0}0.583  & 0.083                          & \cellcolor[HTML]{C0C0C0}0.613  & 0.173                          & \cellcolor[HTML]{C0C0C0}-0.030 & -0.090                         \\
\cellcolor[HTML]{C0C0C0}0.675  & -0.403                         & \cellcolor[HTML]{C0C0C0}0.683  & -0.395                         & \cellcolor[HTML]{C0C0C0}-0.008 & -0.008                         & \cellcolor[HTML]{C0C0C0}0.777  & 0.105                          & \cellcolor[HTML]{C0C0C0}0.744  & 0.204                          & \cellcolor[HTML]{C0C0C0}0.033  & -0.099                         \\
\cellcolor[HTML]{C0C0C0}0.845  & -0.502                         & \cellcolor[HTML]{C0C0C0}0.872  & -0.500                         & \cellcolor[HTML]{C0C0C0}-0.027 & -0.002                         & \cellcolor[HTML]{C0C0C0}0.875  & 0.118                          & \cellcolor[HTML]{C0C0C0}0.803  & 0.230                          & \cellcolor[HTML]{C0C0C0}0.072  & -0.112                         \\ \thickhline
\multicolumn{6}{|c|}{Inward loading case II}                                                                                                                                                          & \multicolumn{6}{c|}{Outward loading case II}\\
\multicolumn{6}{|c|}{$\theta = 30^{\circ}, \delta = 0^{\circ}$}                                                                                                                                     & \multicolumn{6}{c|}{$\theta = 30^{\circ}, \delta = 180^{\circ}$}                                                                                                                                                    \\ \hline
\multicolumn{2}{|c|}{Applied Force}                             & \multicolumn{2}{c|}{Estimated Force}                               & \multicolumn{2}{c|}{Error}                                      & \multicolumn{2}{c|}{Applied Force}                              & \multicolumn{2}{c|}{Estimated Force}                               & \multicolumn{2}{c|}{Error}                                      \\ 
\multicolumn{1}{c}{\textbf{x}} & \multicolumn{1}{c}{\textbf{z}} & \multicolumn{1}{c}{\textbf{x}} & \multicolumn{1}{c}{\textbf{z}} & \multicolumn{1}{c}{\textbf{x}} & \multicolumn{1}{c}{\textbf{z}} & \multicolumn{1}{c}{\textbf{x}} & \multicolumn{1}{c}{\textbf{z}} & \multicolumn{1}{c}{\textbf{x}} & \multicolumn{1}{c}{\textbf{z}} & \multicolumn{1}{c}{\textbf{x}} & \multicolumn{1}{c}{\textbf{z}} \\\hline
\cellcolor[HTML]{C0C0C0}0.180  & -0.078                         & \cellcolor[HTML]{C0C0C0}0.119  & -0.059                         & \cellcolor[HTML]{C0C0C0}0.061  & -0.019                         & \cellcolor[HTML]{C0C0C0}0.195  & 0.024                          & \cellcolor[HTML]{C0C0C0}0.110  & 0.012                          & \cellcolor[HTML]{C0C0C0}0.085  & 0.012                          \\
\cellcolor[HTML]{C0C0C0}0.358  & -0.160                         & \cellcolor[HTML]{C0C0C0}0.228  & -0.114                         & \cellcolor[HTML]{C0C0C0}0.130  & -0.046                         & \cellcolor[HTML]{C0C0C0}0.389  & 0.047                          & \cellcolor[HTML]{C0C0C0}0.303  & 0.075                          & \cellcolor[HTML]{C0C0C0}0.086  & -0.028                         \\
\cellcolor[HTML]{C0C0C0}0.537  & -0.241                         & \cellcolor[HTML]{C0C0C0}0.554  & -0.279                         & \cellcolor[HTML]{C0C0C0}-0.017 & 0.038                          & \cellcolor[HTML]{C0C0C0}0.583  & 0.078                          & \cellcolor[HTML]{C0C0C0}0.524  & 0.112                          & \cellcolor[HTML]{C0C0C0}0.059  & -0.034                         \\
\cellcolor[HTML]{C0C0C0}0.709  & -0.336                         & \cellcolor[HTML]{C0C0C0}0.664  & -0.334                         & \cellcolor[HTML]{C0C0C0}0.045  & -0.002                         & \cellcolor[HTML]{C0C0C0}0.775  & 0.120                          & \cellcolor[HTML]{C0C0C0}0.696  & 0.245                          & \cellcolor[HTML]{C0C0C0}0.079  & -0.125                         \\
\cellcolor[HTML]{C0C0C0}0.887  & -0.419                         & \cellcolor[HTML]{C0C0C0}0.880  & -0.442                         & \cellcolor[HTML]{C0C0C0}0.007  & 0.023                          & \cellcolor[HTML]{C0C0C0}0.871  & 0.142                          & \cellcolor[HTML]{C0C0C0}0.812  & 0.292                          & \cellcolor[HTML]{C0C0C0}0.059  & -0.150                         \\
\thickhline
\end{tabular}
\label{tab:IMU results}
\end{table*}

%% file: content/conclusion.tex
\section{Conclusion}
\label{section:conclusion}

This paper presents the design, modeling, and control of a lightweight modular continuum manipulator, which is aimed to be used in aerial manipulation tasks. Preliminary experimental validations are reported, showing promising potential for future applications and field deployment, including perching and aerial grasping.  
The analytical close-form stiffness formulation and an IMU sensor integration lead to an IMU-based force estimation. The estimation method was developed and experimentally validated.
Future work includes in-flight field deployment tests and improvement on the IMU-based force estimation method.

\section*{Acknowledgment}
This research was supported in part by USDA-NIFA (Grant No. 2021-67022-35977).